\documentclass[letterpaper, 10 pt, conference]{ieeeconf}  

\IEEEoverridecommandlockouts                              

\usepackage{graphicx}
\usepackage{array}
\usepackage{amsmath} 
\usepackage{cite}
\usepackage{color}
\usepackage{booktabs}
\usepackage{makecell}
\usepackage{lipsum}
\usepackage{tabularx}
\usepackage{xcolor}
\usepackage{url}
\usepackage{soul}

\newcommand{\pdce}{\ensuremath{\textrm{pDCE}}}

\newcommand{\cp}[1]{\ensuremath{\textrm{CP}_{\text{#1}}}}
\newcommand{\cc}[1]{\ensuremath{\textrm{CC}_{\text{#1}}}}

\title{\LARGE \bf Responsibility and Engagement --\\ Evaluating Interactions  in Social Robot Navigation}
\author{Malte Probst$^{*}$, Raphael Wenzel$^{*}$, Monica Dasi$^{*}$
\thanks{$^{*}$Honda Research Institute Europe GmbH, Carl-Legien-Str. 30, 63073 Offenbach, Germany, Email:{\tt\footnotesize \{firstname.lastname\}@honda-ri.de}}%
}
\begin{document}

\maketitle
\thispagestyle{empty}
\pagestyle{empty}
\begin{abstract}
In Social Robot Navigation (SRN), the availability of meaningful metrics is crucial for evaluating trajectories from human-robot interactions. In the SRN context, such interactions often relate to resolving conflicts between two or more agents. Correspondingly, the shares to which agents contribute to the resolution of such conflicts are important.
This paper builds on recent work, which proposed a Responsibility metric capturing such shares. We extend this framework in two directions: First, we model the conflict buildup phase by introducing a time normalization. Second, we propose the related Engagement metric, which captures how the agents' actions intensify a conflict. In a comprehensive series of simulated scenarios with dyadic, group and crowd interactions, we show that the metrics carry meaningful information about the cooperative resolution of conflicts in interactions. They can be used to assess behavior quality and foresightedness. We extensively discuss applicability, design choices and limitations of the proposed metrics.
\end{abstract}

\section{Introduction}\label{sec_prob}\noindent

\noindent One goal of Social Robot Navigation (SRN) research is to enable robots to seamlessly navigate complex, unstructured, and potentially crowded environments. Interactions between robots and humans are at the core of SRN. Studies show that humans interacting with publicly deployed robots often compensate for the robots’ lack of navigation capabilities, thereby taking a major share of preventing collisions and conflicting situation \cite{babel_findings_2022, reeves_opening_2025}. While this might be acceptable for the occasional robot, it could turn out to be a nuisance once more robots are deployed in public.
Therefore, research on socially acceptable planning algorithms as well as suitable evaluation metrics has become increasingly important \cite{wang_metrics_2022, gao_evaluation_2022, francis_principles_2025}.
To assess the quality of a navigation algorithm, a meaningful quantitative evaluation of the output trajectories is paramount. Such metrics can also be part of target functions for learning-based approaches.
In this work, we focus on metrics which evaluate various aspects of social compliance of motion behavior, i.e., trajectories.
In \cite{wenzel_spotting_2025}, we pointed out that current metrics fail to address the causal component in interactions between two agents. The degree to which an agent contributed to resolving the conflict is vital for evaluating the quality of said agent's behavior. To this end, we proposed a "Responsibility" metric aiming to quantify the share that agents contributed to de-escalating a potential conflict.

This paper contributes to this line of research. First, we extend the Responsibility ($R$) metric, which is focused on reducing an already existing conflict, to also capture the phase where the conflict is building up over time. Second, we propose the related Engagement ($E$) metric, which quantifies how much the agents' actions are intensifying the conflict.
We then investigate the following questions:

\begin{enumerate}
\item{Do the proposed metrics $R$ and $E$ robustly capture which agent caused an escalation or reduction of a conflict in an interaction?}
\item{Do they help to directly compare the quality of alternative trajectories in a situation?}
\item{Can we use them  to evaluate the quality of planning algorithms in crowd navigation scenarios with higher density and unstructured interactions?}
\end{enumerate}

Establishing meaningful metrics for interactions has the potential to accelerate SRN research: It enables researchers to quantify the effect of a motion planning algorithm regarding the escalation or de-escalation of a conflict in an interaction. For example, it allows to judge if a new parametrization, which makes a robot reach its goal faster, requires the interaction partners to be more accommodating. Moreover, such metrics can be used as part of target functions in algorithm development, both for model-based approaches as well as for data-driven, learning based methods.


%

To address these questions, we conduct a series of simulation experiments with dyadic interactions in different geometric constellations (frontal approach, crossing, and overtaking), scenarios with groups of three agents, and a randomized crowd navigation scenario.


We extensively discuss the metrics, their normative interpretability and the need for grounding, as well as current design choices and limitations.





\section{Related work}\noindent
In recent work \cite{wenzel_spotting_2025}, we showed that frequently used existing metrics do not capture important aspects of two moving agents interacting with each other. Specifically, we analyzed kinematics-based metrics (e.g. an agent's maximum acceleration or mean jerk), metrics based on the distance between two agents (e.g., Clearing Distance or Space Compliance Rate) and prediction-based metrics like Minimum Time-To-Collision or Projected Path Duration \cite{lee_visual_1977, wang_metrics_2022} on a simple scenario: two agents of varying capabilities (oblivious or foresighted) are facing each other on a head-on collision course. While all of the established metrics show meaningful information, they fail to fully highlight a key information: which of the agents (if any) were causally responsible for avoiding the conflict. \cite{wenzel_spotting_2025} propose the Responsibility metric, which is based on predicted minimal distances. In the presented scenarios, the Responsibility metric shows which agent's behavior contributed to the relaxation of the conflict. However, the set of tested scenarios is limited, and all scenarios start with both agents being at full conflict already. Furthermore, no scenarios with more than two agents are addressed.

Other work has also addressed the topic of responsibility in the context of navigation explicitly. \cite{george_feasible_2023} captures causal responsibility of moving agents in their proposed FeAR metric in a discrete, grid-based representation, also assuming that attribution of responsibility should be related to actions (or their lack of). 
Other approaches from the automotive domain also capture or estimate in-situ responsibility, often relying on strong geometrical priors such as lanes or regions \cite{remy2025learning,shalev2017formal}.
Here, we focus on the continuous case, in mostly unstructured environments. We strive for a simple metric for ex-post trajectory evaluation.
Note that the concept of \textit{responsibility} itself is widely used in many different contexts (philosophical, legal, technical, \dots), but, at the same time, inherently ambiguous \cite{vincent2011structured}. This paper focuses on \textit{outcome} or \textit{causal} responsibility responsibility in \cite{vincent2011structured}'s taxonomy.
\section{Responsibility and Engagement Metrics}\label{section:metrics}\noindent
This section extends the Responsibility metric presented in \cite{wenzel_spotting_2025}. First, we revisit the calculation of the underlying conflict potential in Section \ref{section:conflict-potential}. We then propose to derive the \textit{conflict} measure by considering the previously unaddressed conflict buildup phase using a normalization step in Section \ref{section:conflict}. Section \ref{section:conflict-change} explains how the agents' actions are taken into consideration for the escalation or de-escalation of the conflict, before Section \ref{section:responsibility-engagement} shows the updated Responsibility metric and introduces the Engagement metric. Program code for calculating the Responsibility and Engagement metrics from trajectories can be found in a GitHub repository\footnote{\url{https://github.com/HRI-EU/hri-metrics}}.


%

\subsection{Conflict Potential}\label{section:conflict-potential}
\begin{figure}
	\centering
	\includegraphics[width=0.625\columnwidth]{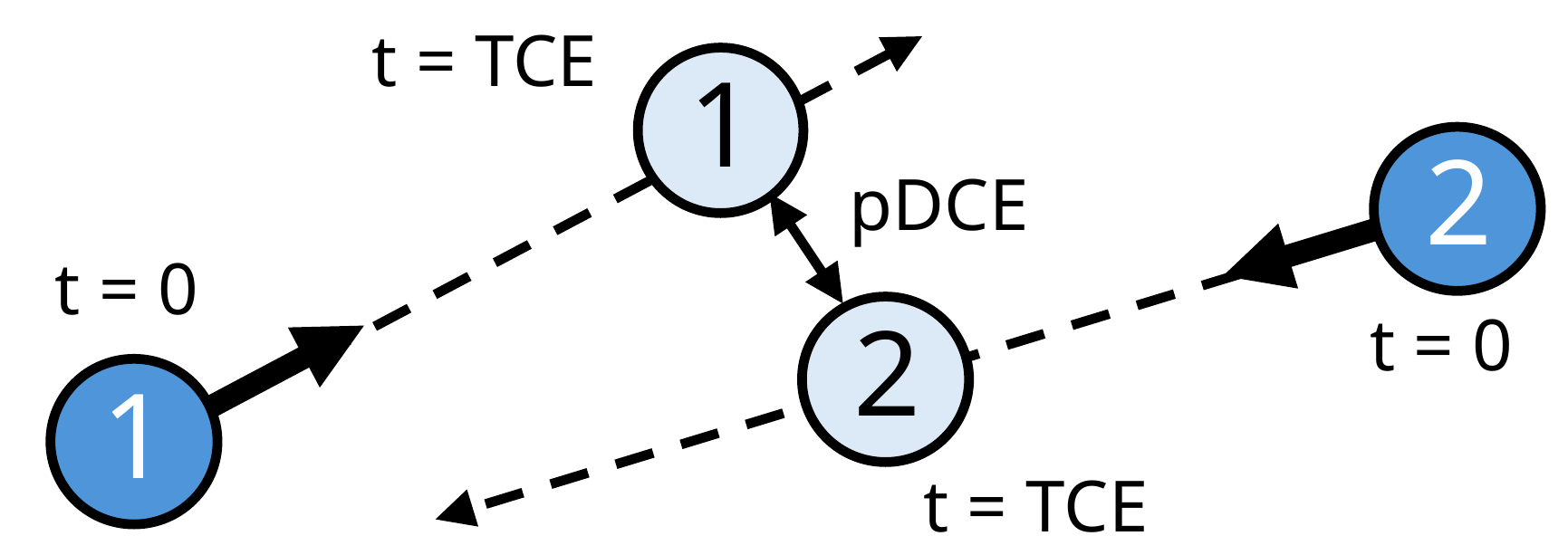}
	\caption{Construction of the predicted Distance of Closest Encounter ($\pdce$). Based on their relative position and velocity, the time to closest encounter (TCE) can be computed. The distance at that time is the $\pdce$.}
	
	\label{fig_dce}
\vspace{-0.5cm}
\end{figure}

\noindent
We use the definition of \cite{wenzel_spotting_2025} to calculate the \textit{Conflict Potential} between two agents (referred to as agent 1 and agent 2) in an interaction. Here, the Distance of Closest Encounter (DCE) \cite{eggert_predictive_2014} of two trajectories plays a central role. To calculate the Conflict Potential, a \textit{predictive} DCE (\pdce) is used, which captures the predicted proximity of two agents continuing along their current trajectories with constant velocity. Figure \ref{fig_dce} shows how the \pdce{} captures the minimum distance between two agents when their movements are predicted based on a constant velocity assumption.


Geometrically, the \pdce{} equals the predicted perpendicular distance, as
\begin{equation}
\label{eq:pdce}
	{\pdce}(t) = \frac{| \mathbf{r}_t \times \mathbf{v}_t |}{|\mathbf{v}_t|}. 
\end{equation}
%
\noindent Here, $\mathbf{r}_t$ is the relative position vector and $\mathbf{v}_t$ the relative velocity vector of both agents at time $t$.


Based on the $\pdce$, the \textit{Conflict Potential} $\cp{}$ of a situation is given as
%
%
\begin{equation}
\label{eq:cp}
	{\cp{}}(t) =\max(0, 1 - \frac{{\pdce}(t)}{r_{\text{1}} + r_{\text{2}}}),
\end{equation}

\noindent where $r_{\text{1}}$ and $r_{\text{2}}$ are the radii of agents 1 and 2, assuming circular agents.
In a nutshell, the conflict potential equals one if the agents are on a direct collision course. It decreases as their predicted center positions at the point of closest encounter move further apart, becoming zero when they are predicted to just miss each other narrowly.

\subsection{Conflict}\label{section:conflict}\noindent
The original concept of conflict potential does not take the conflict buildup phase into account. Two agents, which are 100m apart, walking straight towards each other have a conflict potential of $\cp{}(0)=1$.
While technically correct, we need to capture the fact that a potential conflict does not necessarily lead to an immediate reaction by the agents, e.g. for collision avoidance \cite{mirsky_conflict_2024}.


We use the time of closest encounter (TCE) as an anchor point for the interaction. At times $t > \text{TCE}$, agents not at standstill are moving apart from each other, so the most important part of the interaction is when $t < \text{TCE}$.
Since the agents' actions long before the TCE ($t\ll\text{TCE}$) are typically not influenced by the interaction, we introduce a time-dependent, linear normalization term $N(t)$ as
\begin{equation}
\label{eq:normalization}
	N(t) = 
	\begin{cases}
		\frac{1}{N_W}t + 1 - \frac{\text{TCE}}{N_W}& \text{if } \text{TCE} - N_W \leq t \leq \text{TCE}\\
		0 & \text{else, }
	\end{cases}
\end{equation}
\noindent where $N_W$ is the length of the weighted window.
With this understanding, the Conflict C can be defined as
\begin{equation}
\label{eq:conflict}
	C(t) = \cp{}(t) N(t).
\end{equation}

\begin{figure}
\begin{centering}
	\includegraphics[trim={0.5cm 0 0.2cm 0},clip,width=0.72\columnwidth]{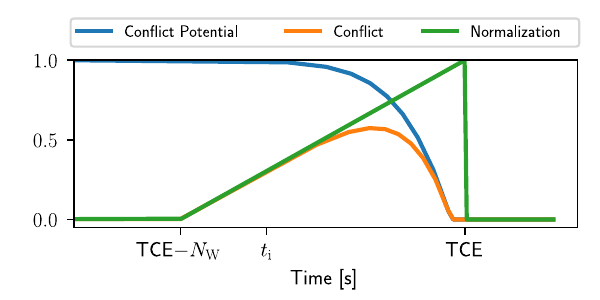}
\vspace{-0.2cm}

	\caption{Exemplary depiction of the relationship between Potential Conflict, Normalization, and resulting Conflict for a generic interaction scenario where two agents are on a collision course (c.f. Experiment 1, Oncoming, Scenarios 2 and 3). The Conflict Potential is high from the beginning, but the normalized Conflict starts building up only from $t=\text{TCE}-N_W$ onward. At $t>t_i$, one agent gradually changes its course to avoid the collision, causing the Conflict Potential to decrease. Accordingly, the Conflict buildup slows down, and, eventually, the Conflict also starts to decrease, until both Conflict Potential and Conflict reach zero shortly before $t=\text{TCE}$.}
	\label{fig:c-normalization-cp}
\end{centering}
\vspace{-0.5cm}
\end{figure}

Figure \ref{fig:c-normalization-cp} illustrates the relationship between conflict potential, the time normalization and the resulting conflict measure. See Section \ref{section:discussion} for a discussion on alternatives to using the TCE and the choice of the window size $N_W$.

\subsection{Agent Behavior and Conflict Contribution}\label{section:conflict-change}\noindent
We follow the notion of \cite{wenzel_spotting_2025} and \cite{george_feasible_2023} that an agent's responsibility in an interaction can be derived by calculating how much it contributed to the conflict by changing its trajectory, or, more generally put, its behavior $\mathcal{B}$.
To calculate an agent's interaction-related change of action $\dot{\mathcal{B}}^\text{int}$, we first need to define its \textit{reference} behavior $\mathcal{B}^\text{ref}$, i.e., the behavior the agent would have displayed in case there were no interaction.

Given perfect information about the agent (such as its goals, kinematics, or preferences) and the environment (such as static objects, or terrain slope), $\mathcal{B}^\text{ref}$ would be the optimal trajectory towards the agent's goal\footnote{Similarly, with perfect information, the \pdce{} in Eq. \ref{eq:pdce} could be calculated more precisely using geometry- and agent- specific data for $\mathbf{r}_t$ and $\mathbf{v}_t$}. However, such information is typically not available.
For the remainder of this paper, we therefore assume that any agent's reference behavior is to continue its current behavior $\mathcal{B}$ with constant velocity at any point in time.
Accordingly, the change in reference behavior $\dot{\mathcal{B}}^\text{ref}$ is always zero. Therefore, the interaction-related change $\dot{\mathcal{B}}^\text{int}$ is equal to its absolute change in behavior $\dot{\mathcal{B}}$.


With these assumptions, we can approximate the conflict contribution (CC) of an agent as the change in conflict due to the agent's change in behavior:
\begin{equation}
	\cc{agent}(t) = \frac{d}{d \mathcal{B}_\text{agent}} C(t) \approx C(t) - C_\text{agent, no change}(t).
\end{equation}
\noindent Here, we calculate $C_\text{agent,no change}(t)$ by using the \pdce{} in Eq. \ref{eq:pdce} with the relative velocity vector $\mathbf{v}_{t}$ based on one agent's previous absolute velocity $\mathbf{v}_{\text{abs}, t-1}$ and the other agent's current absolute velocity $\mathbf{v}_{\text{abs}, t}$ to estimate the contribution of the agent's current velocity change on the overall conflict.

To further investigate how the agents contributed to the evolution of the conflict, we split the Conflict Contribution for each agent in its contribution to escalate the conflict, 
\begin{equation}
	CC_\text{agent}^+(t) = \max{(\cc{agent}(t), 0)}
\end{equation}

\noindent and its contribution to diminishing the current conflict
\begin{equation}
\label{eq:cc-}
	CC_\text{agent}^-(t) = - \min{(\cc{agent}(t), 0)}.
\end{equation}








If two agents walk towards each other, the conflict increases over time according to Eq. \ref{eq:conflict}.
However, if both agents are merely following their reference behavior, it is unclear who is to blame for this increase.
Therefore, we postulate that the reference behavior is neither good nor bad and each agent has a right to follow its respective reference behavior and the conflict escalates "over time". This interpretation is in line with similar ideas in \cite{george_feasible_2023}.


We define the time's escalating contribution $CC_\text{Time}^+$ to the conflict by subtracting all agents' contributions to the conflict from the overall change $\dot{C}(t)$ in conflict
\begin{equation}
	CC_\text{Time}^+(t) := \max{(\dot{C}(t) - \sum_\text{agents} CC_\text{agent}^+(t), 0)}.
\end{equation}

\noindent Vice versa, we define the time's diminishing contribution as
\begin{equation}
	CC_\text{Time}^-(t) := - \min{(\dot{C}(t) - \sum_\text{agents} CC_\text{agent}^-(t),0)}.
\end{equation}

One interpretation is that $CC_\text{Time}^-$ is the residual conflict which was not fully resolved by the agents.
As the distance between agents increases again, any remaining conflict is therefore diminished by time. 

\subsection{Responsibility and Engagement}\label{section:responsibility-engagement}\noindent
Based on the contributing factors for agents and the time, we can determine the total conflict $C_{total}$ which existed over the course of the interaction:
\begin{multline}
	\label{eq_resp}
	C_\text{total} = \int (\sum_\text{agents} CC_\text{agent}^+(t) + CC_\text{Time}^+(t))\,dt.
\end{multline}


Now, we can define the Responsibility assigned to $x \in [\text{Agent 1, Agent 2, Time}]$ in an interaction by determining how much the agent contributed to diminishing the conflict relative to the overall total conflict:
\begin{equation}
\label{eq_resp}
R_x = \frac{1}{C_\text{total}}\int CC_x^-(t) \,dt, \\
\end{equation}
$$
\text{where} \sum_{x}{R_x} = 1.
$$
Likewise, we can define the Engagement E of an agent in an interaction as the contribution to escalating or intensifying the conflict.
%
%
\begin{equation}
	\label{eq_resp}
	E_x = \frac{1}{C_\text{total}}\int CC_x^+(t) \,dt,
\end{equation}
$$
\text{where} \sum_{x}{E_x} = 1.
$$
Note that this escalation is not necessarily caused by malicious intent of an agent. It could be, for example, the result of another, previous interaction (agent 1 avoiding agent 2, thereby engaging agent 3, which is farther away), or a deliberate approach (agent 1 wants to start a conversation with agent 2).

\section{Experiments}\label{sec_exp}\noindent
\subsection{Experimental Setup and Notation}\noindent
We test the metrics in simulations of interactive scenarios using different agent types. The most simple type are "ballistic", unresponsive agents, which move on a fixed heading and with constant velocity, oblivious of their surroundings. On the other hand, interactive, responsive agents are controlled by a Social Force model \cite{helbing_social_1995}\footnote{Parameters: $A$~=~5.1, $\lambda~=~3.0$, $\gamma$~=~0.35, $n$~=~1 and $n'$~=~3.0)}.
We set $r_\text{1}=r_\text{2}=0.5\text{ m}$. 
In order to be conservative regarding the length of the full conflict built-up phase, we set the time normalization window size to $N_W=12$ s.

For easier reference in Experiments 1, 2, and 4, we refer to the first agent as the \textit{Robot}, the second agent as \textit{Alice} and the third agent as \textit{Bob}. Whenever we refer to the Responsibility of Alice in the interaction between Alice and Bob, we note $R^\text{Alice, Bob}_\text{Alice}$. In case the pair of agents is clear from context, we omit the interaction partners and note $R_\text{Alice}$ for conciseness.

\subsection{Experiment 1: Dyadic Interactions}\noindent
We analyze how the metrics capture dyadic interactions between two agents in various geometric constellations. Specifically, we simulate scenarios where agents are walking straight towards each other, crossing each other at 90°, and overtaking each other. For all geometric constellations, we investigate four different cases: Both agents are unresponsive, only the robot is responsive, only Alice is responsive, or both the robot and Alice are equally responsive.

First, we repeat the scenario from \cite{wenzel_spotting_2025} to verify that the proposed extensions to the Responsibility metric still yield the same intuitive results. Here, the robot is on a head-on collision course with Alice. 
\begin{figure}
	\includegraphics[width=\columnwidth,trim=0 0.6cm 0 0]{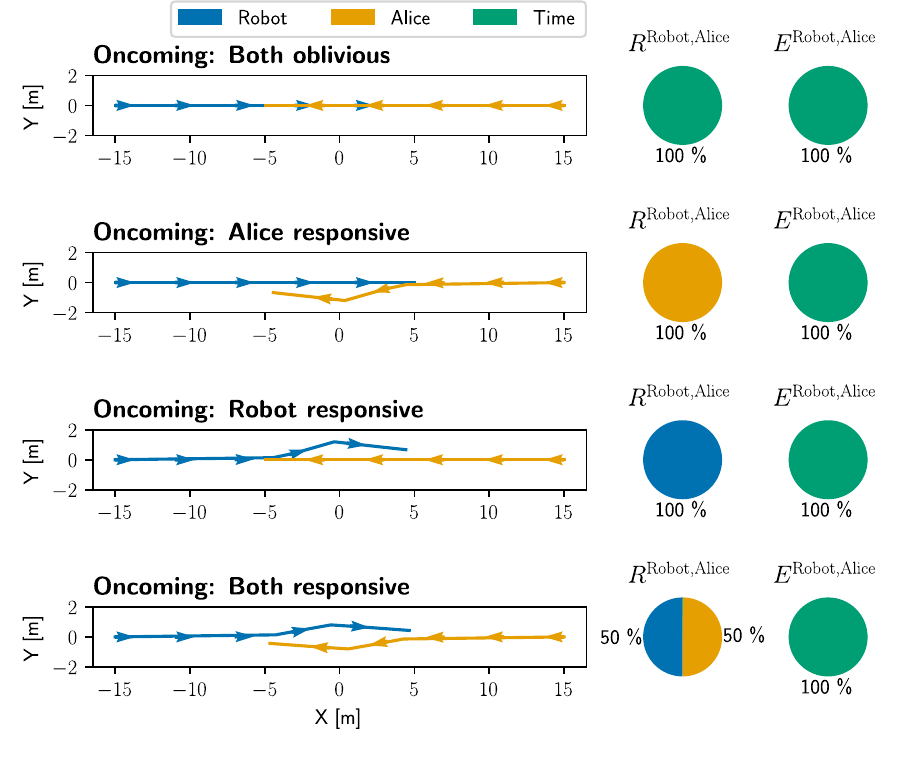}
	\caption{Experiment 1 - Oncoming scenarios; Trajectories and Responsibility/ Engagement metrics for all scenarios. Arrow sizes indicate velocities, sampled at equidistant time intervals.}
	\label{fig:oncoming}
\vspace{-0.4cm}
\end{figure}
Figure \ref{fig:oncoming} shows the trajectories as well as Responsibility and Engagement for the four cases.
In Scenario 1, neither agent takes over Responsibility for the resolution ($R_\text{Robot}=R_\text{Alice}=0\%$), the resulting conflict is "resolved" by time, i.e., the agents collide ($R_\text{Time}=100\%$  ). In Scenario 2 and 3, one agent evades, thereby taking over all Responsibility for avoiding the collision ($R_\text{Alice}=100\%$ in Scenario 2, $R_\text{Robot}=100\%$ in Scenario 3). In Scenario 4, both agents are simulated based on Social Force and therefore behave symmetrically. In this case, the Responsibility of avoiding a collision distributed equally on both agents, as indicated by $R_\text{Robot}=R_\text{Alice}=50\%$.
In all four scenarios, no agent actively increases the conflict by changing its actions. Therefore, the Engagement in all scenarios is identical ($E_\text{Time}=100\%$).

\begin{figure}
	\includegraphics[width=\columnwidth]{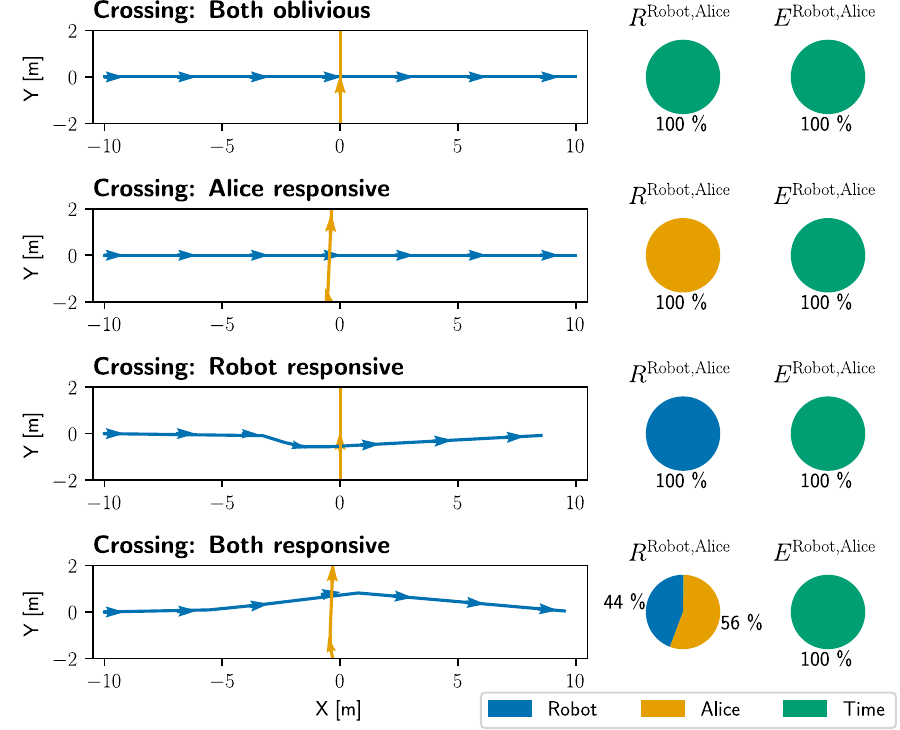}
	\caption{Experiment 1 - Crossing scenarios}
	\label{fig:crossing}
\end{figure}

Second, we simulate scenarios where the robot and Alice are walking towards each other at 90° angle (see Figure \ref{fig:crossing}). Identical to the Oncoming constellation, neither agent takes Responsibility, and they crash in Scenario 1 ($R_\text{Robot}=R_\text{Alice}=0\%$). In both Scenario 2 and 3, the reactive agent takes Responsibility by slightly evading and, simultaneously, reducing its velocity (indicated by arrow sizes). Again, this is captured by the metric, assigning full Responsibility to Alice in Scenario 2 ($R_\text{Robot}=0\%$, $R_\text{Alice}=100\%$, and the Robot in Scenario 3 ($R_\text{Robot}=100\%$, $R_\text{Alice}=0\%$).
In the scenario where both the robot and Alice are controlled by the Social Force algorithm, their respective evasion behaviors differ slightly (not visible in the plot). The robot evades mostly laterally, with a maximum deviation of $83 \text{ cm}$. Alice's deviates laterally with a maximum offset of $43 \text{ cm}$, but simultaneously reduces her velocity to $0.75 \frac{\text{m}}{\text{s}}$. The Responsibility metric shows that both agents contributed, but assigns a slightly bigger share to Alice ($R_\text{Robot}=44\%$, $R_\text{Alice}=56\%$). Again, the Engagement scores are zero for both agents in all scenarios.

\begin{figure}[ht!]	 
	\includegraphics[width=\columnwidth]{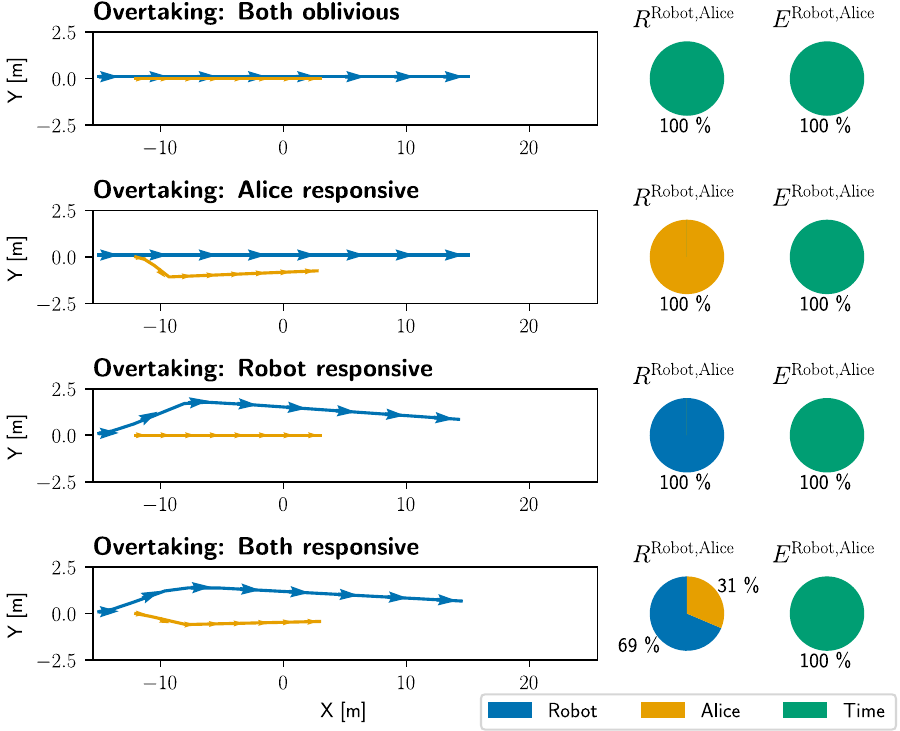}
	\caption{Experiment 1 - Overtaking scenarios}
	\label{fig:overtaking}
\vspace{-0.3cm}
\end{figure}

Lastly, we simulate a situation where Alice, walking at a leisurely pace of $v_{\text{A}}=0.5\frac{\text{m}}{\text{s}}$, is overtaken by the robot moving forward at $v_{\text{R}}=1.0 \frac{\text{m}}{\text{s}}$. Figure \ref{fig:overtaking} shows the paths of both agents and the resulting Responsibility and Engagement.
Once again, neither agent evades or takes Responsibility in Scenario 1 ($R_\text{Robot}=R_\text{Alice}=0\%$), leading to a crash. In Scenario 2, Alice takes full Responsibility and jumps out of the way of the oblivious robot approaching from behind ($R_\text{Robot}=0\%$, $R_\text{Alice}=100\%$), and the robot evades by going around the slower Alice in Scenario~3 ($R_\text{Robot}=100\%$, $R_\text{Alice}=0\%$). In the last scenario where both agents are responsive, they both evade, but since the robot is moving and turning faster due to its higher pace, its Responsibility evaluates higher than Alice's ($R_\text{Robot}=64\%$, $R_\text{Alice}=35\%$). As before, the corresponding Engagement scores for the robot and Alice are zero.

This experiment provides two insights: First, since the Responsibility metric is based on the \pdce{}, it covers diverse interactions, regardless of the orientation of both agents toward each other. Also, it captures both longitudinal and lateral behavior changes.
Second, while Responsibility is a valuable metric to investigate behavior in interactions, it does not entail any direct normative interpretation. 
In order to evaluate if a behavior was acceptable, the context is vital. Consider Scenarios 2 and 3, where only one agent is responsive, for the Oncoming and Overtaking case. Normatively, we would expect an agent to be alert, taking others into account when planning its trajectory. Hence, a Responsibility of zero typically corresponds to suboptimal (or at least unfriendly) behavior. 
However, in Scenario 3 of the Overtaking geometry, Alice's oblivious behavior is fully acceptable, since she can't be expected to see the robot approaching from behind. Vice versa, in Scenario 2, the robot is threatening to run over Alice from behind - which is completely unacceptable, since Alice has to jump away to save herself.
Similarly, in the Overtaking case, Alice's Engagement would rise if she slowed down while the robot is approaching from behind (since she actively changes her behavior). 



\subsection{Experiment 2: Group splitting}
\begin{figure}
	\includegraphics[width=0.5\textwidth]{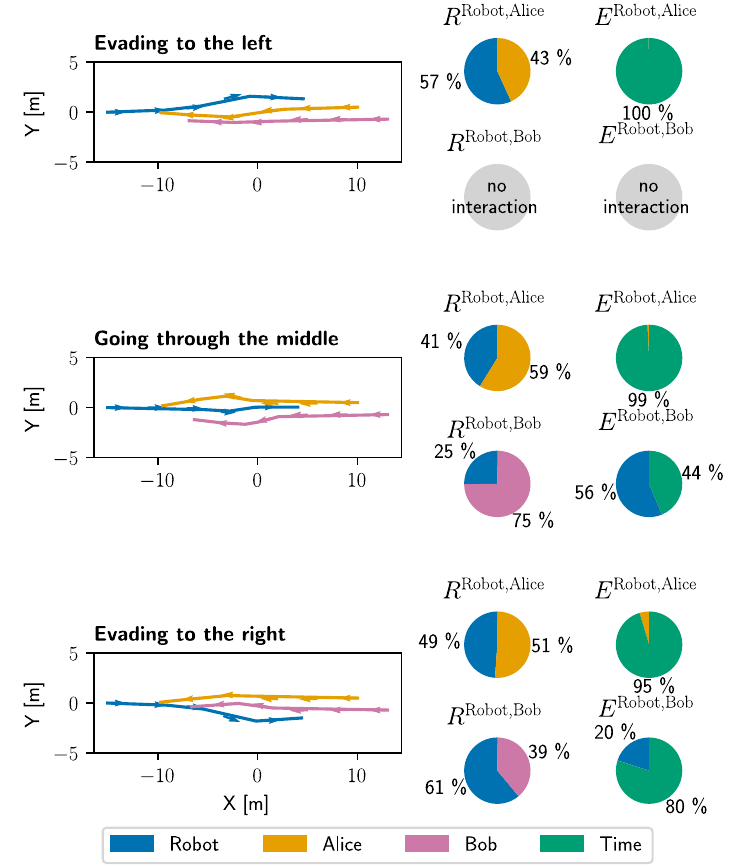}
	\caption{Experiment 2 - Group Splitting scenarios}
	\label{fig:group-splitting}
\vspace{-0.3cm}
\end{figure}

\noindent
Next, we investigate the interaction of a robot with a group of two people, Alice and Bob. All agents are interactive, i.e. controlled by the Social Force model. Alice and Bob walk towards the robot, with an initial distance of 25~m. Alice is positioned slightly in front of Bob and towards the left-hand side of the robot.
We now run three variations of the scenario, by slightly shifting the robot's goal to be behind Alice (Scenario 1), behind but in between Alice and Bob (Scenario 2), or behind Bob (Scenario 3). 
This slight shift creates three semantically different behaviors: In Scenario 1, the robot passes both Alice and Bob on its left side. In Scenario 2, the robot forces its way in between the two, splitting Alice and Bob. In Scenario 3, the robot passes both agents on its right side. The results for all three scenarios are depicted in Figure \ref{fig:group-splitting}.

The first scenario shows the robot evading the group towards its left side. Here, the Responsibility metric indicates only a direct interaction with Alice, in which both the robot and Alice take over the resolution of the conflict. As Alice's trajectory was also influenced by Bob, who is walking behind her, the robot takes over slightly more Responsibility ($R_\text{Robot}^{\text{Robot,Alice}}=57\%$ vs $R_\text{Alice}^{\text{Robot,Alice}}=43\%$). Neither the robot nor Alice actively intensify the mutual conflict, hence the Engagement comes from time ($E_\text{Time}^{\text{Robot,Alice}}=100\%$).

By placing the goal straight behind the humans, the robot moves directly in between Alice and Bob in the second scenario, interacting with both agents simultaneously. Correspondingly, it takes over less Responsibility compared to Alice ($R_\text{Robot}^{\text{Robot,Alice}}=41\%$, $R_\text{Alice}^{\text{Robot,Alice}}=59\%$). Similarly, the robot takes less Responsibility than Bob in their interaction ($R_\text{Robot}^{\text{Robot,Bob}}=25\%$, $R_\text{Bob}^{\text{Robot,Bob}}=75\%$), since it is constrained by Alice.
Additionally, the Engagement metric reveals that, while evading Alice, 
the robot intensified the conflict with Bob ($E_\text{Robot}^{\text{Robot,Bob}}=56\%$). 


In the third scenario, the robot chooses to evade both Alice and Bob on its right side. 
In the interaction with Alice, the robot swerves decisively to the right, and the Responsibility share between them is balanced ($R_\text{Robot}^{\text{Robot,Alice}}=49\%$, $R_\text{Alice}^{\text{Robot,Alice}}=51\%$). 
Here, the robot decides to evade Bob on its right side as well. Since the underlying Social Force algorithm is not constrained by Alice's trajectory, it can evade Bob in a wider arc as well. Accordingly, the robot takes over more Responsibility with respect to Bob, who is now constrained by Alice himself ($R_\text{Robot}^{\text{Robot,Bob}}=61\%$, $R_\text{Bob}^{\text{Robot,Bob}}=39\%$). This also results in a lower Engagement with Bob while evading Alice, compared to Scenario 2  ($E_\text{Robot}^{\text{Robot,Bob}}=20\%$).

The three scenarios show that different ways of passing Alice and Bob (induced by slightly different goals) lead to different levels of Responsibility and Engagement of the robot.
The subjectively best strategy (going left of Alice and Bob, avoiding the conflict with Bob altogether) gives the highest mean Responsibility, and no Engagement. The subjectively second-best strategy (avoiding Alice and Bob to the right, interacting with both) gives the second highest mean Responsibility and some Engagement. The strategy with the highest perturbation (going straight through the oncoming group) results in the lowest mean Responsibility, and highest mean Engagement, which seems plausible, since both Alice and Bob evade the robot actively. 
\subsection{Experiment 3: Crowd Navigation}\noindent
Now, we focus on a scenario where the robot interacts with many different other agents. We simulate a rectangular 10m by 10m area that the robot needs to traverse. Within the area, 20 human agents start at random locations (average density of $0.2\frac{\text{agents}}{\text{m}^2}$). Controlled by the Social Force algorithm, they navigate to random goals on the edges of the area, which are resampled on arrival. All agents' target velocity is $1.0 \frac{\text{m}}{\text{s}}$.

We run three different variations of the scenario: First, the robot is simulated as a ballistic agent, oblivious of its surroundings.
Second, the robot is controlled by the Dynamic Window Approach (DWA)\cite{fox_dynamic_1997}, but constrained to a sensor range of 1~m. This shortsighted agent will head for its goal, and only deviate (evade or slow down) on imminent collisions.
Third, the robot also uses the Social Force algorithm.
For each case, we run 50 repetitions, with fixed random seeds across robot types.

\begin{figure}
	\includegraphics[width=0.5\textwidth]{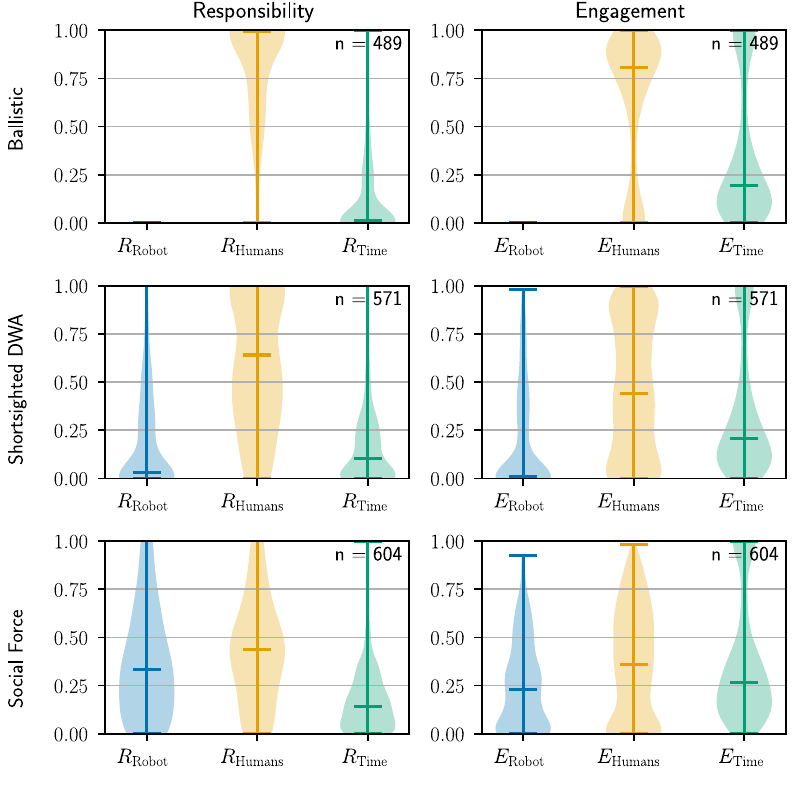}
	\caption{Experiment 3: Crowd Navigation; Distributions for Responsibility and Engagement metrics for different agent types. $n$ indicates the number of interactions for each distribution, horizontal bars are median values.}
	\label{fig:crowd-navigation}
\vspace{-0.3cm}
\end{figure}

Figure \ref{fig:crowd-navigation} shows the distributions of Responsibility and Engagement and the respective median values $\tilde{R},\tilde{E}$ across the three cases.
By definition, the ballistic robot (first row) shows neither Responsibility nor Engagement. Consequently, the human agents take most of the Responsibility ($\tilde{R}_{\text{Humans}}^{\text{Ballistic}}\approx 99\%$), and some cases were resolved "by time", i.e., agents came closer than 1.0m ($\tilde{R}_{\text{Time}}^{\text{Ballistic}}\approx 1\%$).
Also, most Engagement towards the robot comes from the human agents, who need to maneuver in a crowded space (median $\tilde{E}_{\text{Humans}}^{\text{Ballistic}}\approx81\%$).

The distribution of Responsibility of the shortsighted DWA robot (second row) shows that the agent can now actively avoid imminent collisions ($\tilde{R}_{\text{Robot}}^{\text{DWA}}\approx3\%$). Nevertheless, most of the evasions are done by the human agents ($\tilde{R}_{\text{Humans}}^{\text{DWA}}\approx64\%$). At the same time, the collision avoidance actions now lead to a higher Engagement with other agents in some cases ($\tilde{E}_{\text{Robot}}^{\text{DWA}}\approx1\%$).

Lastly, we evaluate the metrics for the robot controlled by Social Force. Compared to the shortsighted DWA agent, it shows a much higher Responsibility ($\tilde{R}_{\text{Robot}}^{\text{SF}}\approx33\%$) and more Engagement ($\tilde{E}_{\text{Robot}}^{\text{SF}}\approx23\%$). The distributions of $R$ and $E$ are similar to those of the human agents, which is to be expected, since they are controlled by the same algorithm.

Together, the results show that a more foresighted algorithm evaluates with a higher Responsibility than a shortsighted, or oblivious agent. At the same time, since the foresighted agents tend to maneuver more, the metrics also attribute more Engagement to their actions.

\subsection{Experiment 4: Playing Catch}\noindent
We now investigate a scenario which is not at the core of Social Robot Navigation, but, nevertheless, interesting for the Responsibility and Engagement metrics. Specifically, we take a look at the Robot and Alice playing catch. Here, the robot is the runner, and Alice plays the chaser.
Alice is controlled by a slightly modified DWA \cite{fox_dynamic_1997} algorithm. She makes a linear prediction of where the robot will be in $l=d/1.2$ seconds, with $d$ being the distance between Alice and the robot. The predicted position is used as the goal for DWA. 
Additionally, DWA is limited by a sensor range of 1~m, i.e., the algorithm is shortsighted w.r.t. collision avoidance. Hence, Alice homes in on the robot's future position, evading only on imminent collisions. For simplicity, the robot is controlled by a standard Social Force algorithm, trying to reach a goal which is far away while keeping its distance to other agents.

\begin{figure}
	\includegraphics[width=0.5\textwidth]{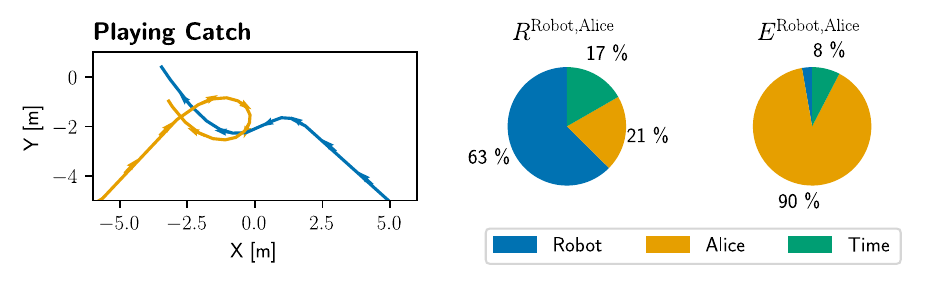}
	\caption{Experiment 4 - Playing Catch scenario, Alice chases the robot}
	\label{fig:playing-catch}
\end{figure}

Figure \ref{fig:playing-catch} shows one interaction of the game. The robot starts on the bottom right corner going diagonally to the upper left corner, with Alice starting in the bottom left corner, tying to intercept its way. The robot reacts by evading to the left. Alice readjusts the course but overshoots, starting to go in a narrow right curve, just avoiding a collision. the robot keeps evading. 
Since this point in time marks the interaction's time of closest encounter (TCE, see Eq. \ref{eq:normalization}), the rest of the interaction (i.e., Alice and the robot completing the circle) is not captured.

Most of the conflict intensification is due to active maneuvering by Alice ($E_\text{Alice}=90\%$). Most of the Responsibility is credited to the robot ($R_\text{Robot}=63\%$), some for Alice who - if shortsightedly - contributed to avoiding the imminent collision ($R_\text{Alice}=21\%$). The metric also shows a relaxation of the conflict by time, marking the residual conflict at the TCE ($R_\text{Time}=17\%$).
\subsection{Experiment 5 - Personal Space Compliance}
\label{section:proxemics}
\begin{figure}
	\includegraphics[width=0.5\textwidth]{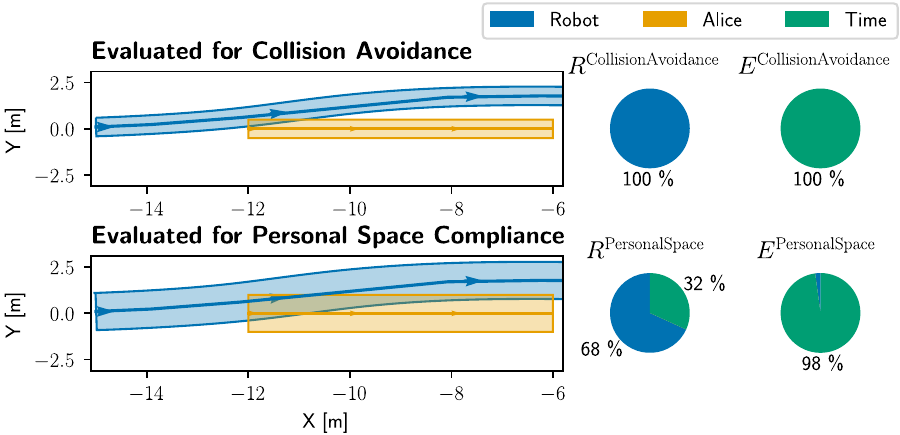}
	\caption{Experiment 5 - Overtaking scenario, evaluated for both Collision Avoidance and Personal Space Compliance. Shaded areas in trajectory plots indicate the width of the conflicting area around the agents.}
	\label{fig:proxemics}
\vspace{-0.3cm}
\end{figure}

In Eq. \ref{eq:cp}, we can choose the agent sizes to determine which predicted distance leads to a potential conflict. Accordingly, it is possible to evaluate $R$ and $E$ with respect to proxemics \cite{hall_hidden_1966} instead of pure collision avoidance.
Here, we reevaluate the case where the robot overtakes Alice, who does not see the robot behind her. This time, we increase the sizes $r_\text{1}$ and $r_\text{2}$ from 0.5~m to 1.0~m. Correspondingly, Eq. \ref{eq:cp} will result in a conflict whenever $\pdce \le 2$~m. Assuming circular agents with a radius of 0.5~m, this amounts to 1~m of free space between the agents, which can roughly be considered the boundary of personal space \cite{hall_hidden_1966}.
Figure \ref{fig:proxemics} shows the resulting Responsibility and Engagement, evaluated with respect to Collision Avoidance and Personal Space Compliance.
By passing close to Alice, the robot avoids a collision, but violates her personal space.
Accordingly, using the increased size, the robot's Responsibility is now lower, with $R_{\text{Robot}}^\text{PersonalSpace}=64\% < R_{\text{Robot}}^\text{Collision Avoidance}100\%$.
Furthermore, there is a residual conflict at the time of closest encounter, which is captured by $R_\text{Time}^\text{PersonalSpace}=36\% > R_\text{Time}^\text{CollisionAvoidance}=0\% $.
Since the robot did not actively turn \textit{towards} Alice after evading the collision, the corresponding Engagement is still zero ($E_{\text{Robot}}^\text{PersonalSpace}=0\%$).

The same pattern applies to the other scenarios where the agents are unresponsive, or controlled by pure collision avoidance algorithms. On increasing the sizes for the metrics evaluation, the Responsibility values for all agents decrease, and $R_\text{Time}$ increases, since there are residual conflicts, i.e., violations of personal space, at the TCE. Correspondingly, the $E$ values typically increase if agents actively increase the personal space conflict before the TCE, e.g., by turning back to face their original goals.


\section{Discussion}\label{section:discussion}\noindent
\vspace{-0.7cm}
\subsection{Research Questions, Interpretability, and Grounding}\label{section:normative}
With respect to our first research question, the dyadic interaction scenarios of Experiments 1 and 4 show that the proposed metrics reliably quantify which agent contribute to escalating or diminishing a conflict, in a variety of different geometric constellations, thereby complementing existing metrics \cite{wenzel_spotting_2025}. Furthermore, Experiment 5 shows that they also capture residual conflicts caused by a collision avoidance algorithm entering the personal space of another agent.

The Responsibility and Engagement metrics do, by definition, not directly entail a normative interpretation of the underlying behavior. 
Experiment 1 also showed that the context of an interaction is crucial for drawing such normative conclusions: While an $R$ value of zero is totally acceptable for an agent which is approached from behind by a faster agent, it is typically not good behavior to show the same, unresponsive behavior in a head-on interaction.

The group splitting scenario of Experiment 2 addresses the second research question. Here, the different trajectory options for the robot share the same evaluation context (e.g., a similar scenario start and progression). In this case, the values of $R$ and $E$ enable a direct comparison of different trajectory alternatives 
(going left, right, or through the group). The ordinal ranking of the $R$ metric matches a subjective ranking of the behavior quality.

The crowd navigation scenario in Experiment 4 focuses on the third research question. Although conflicts are defined on dyadic interactions, collecting distributions of $R$ and $E$ over multiple randomized interactions enables the comparison of different agent types navigating in a crowded, unstructured environment. Again, the ordinal ranking of the average Responsibility is plausible: trajectories generated by more foresighted algorithms evaluate with a higher $R$, on average. Accordingly, the proposed metrics can be a useful tool for analyzing different planning algorithms.

In order to increase the normative interpretability, or even compare different values on an interval level, the metrics need to be grounded with recorded trajectories of human interactions. In future work, we will use trajectory data from public data sets to analyze the observable distributions of $R$ and $E$. After that, running user studies can help to understand which levels of $R$ and $E$ humans would \textit{expect} in interactions with another agent - robot or human.

\subsection{Current limitations and potential extensions}\label{section:limitations}\noindent
The proposed metrics come with a number of design choices and limitations.
First, by construction, they are dyadic in nature, focusing on the interaction conflict between two agents. Therefore, second order effects (while avoiding Alice, the robot is turning towards Bob) are captured only indirectly, without active consideration. Similarly, intricate social norms are not addressed. Nevertheless, aggregated statistics of sequential or simultaneous interactions with multiple agents can help to interpret agent behavior (see crowd navigation experiment). A possible extension to improve those aggregated statistics would be to include a measure of the conflict intensity in the process, e.g. to create weighted averages.

Second, if the behavior of two agents is strongly coupled, the metrics can be skewed. For example, in simulation, we observed the effect that two agents are locked in a local optimum. They walk next to each other, moving perpendicular with respect to their respective goals, blocking each other, and reducing their velocity gradually. In this special case, Eq. \ref{eq:cc-} yields significant diminishing conflict contributions for both agents, since it assumes that only one agent changes its behavior. In contrast, the actual total conflict across two time steps remains nearly constant. Artifacts due to perfect symmetry are less likely to occur in reality, nevertheless they could be a potential source of trouble.

Due to the assumption that the reference behavior is a constant velocity trajectory (see Section \ref{section:conflict-change}), the proposed metrics work best for open space scenarios. It would be worthwhile to relax this assumption, e.g., by considering curvy reference paths by accounting for their angle in the calculation of $d \mathcal{B}$.

Also, the choice of anchor point for the normalization can be reassessed (see Sect. \ref{section:conflict}). The current choice to tie the end of the normalization window to the time of closest encounter (TCE), an early reaction of agent A has a lower effect on $R$ than a late reaction of its counterpart $B$. Taking the actual end of the conflict (or time of lowest conflict) into account could mitigate this artifact.
Similarly, the choice of the window size $N_W$ as well as the disambiguation of consecutive interactions between two partners need more consideration. As before, carefully grounding the metrics in trajectory data from human interactions may help to get a better understanding, e.g., on the influence of crowd density.


Finally, while Responsibility and Engagement show promising results for evaluating the quality of agent behavior, a joint metric which combines the two perspectives could help to give an even more concise evaluation. 

\section{Conclusion}\noindent
A current topic in Social Robot Navigation research is to provide useful evaluation metrics for understanding and assessing the quality of an agent's behavior in interactions. Recent work has shown that many metrics fail to capture which agent contributed to escalating or reducing the conflict for a shared resource, i.e., space.
We continue this line of work by extending the Responsibility metric presented in \cite{wenzel_spotting_2025}, which quantifies agents' contributions towards conflict resolution. The extended Responsibility ($R$) metric now captures the previously neglected phase of conflict buildup. Furthermore, we propose the Engagement ($E)$ metric, which, analogously, captures conflict escalations by agents.


In a series of experiments with simulated dyadic interactions, we showed that $R$ and $E$ reliably detect the agents' contributions to conflicts across different 2D geometries, incorporating variations of speed as well as lateral evasion maneuvers. Specifically, $R$ shows which agent contributed more to the resolution of the conflict by changing its behavior (e.g., by slowing down or evading). Similarly, $E$ highlights if an agent actively intensifies an interaction conflict, e.g., by turning towards another agent.
While $R$ and $E$ do not entail normative interpretations directly, they can make behavior quality comparable when evaluating interactions in a similar context: In the experiments with small groups, the subjective quality of the trajectories matched the ordinal ranking of their $R$ values.
Furthermore, although the metrics are derived from dyadic interactions, we can analyze their distribution over multiple interactions between different interaction partners. The crowd navigation experiment showed that $R$ and $E$ can help to draw conclusions about the foresightedness of the underlying planning algorithm from trajectory data.

By varying the metrics' parameters, we use them to evaluate interactions with respect to collision avoidance or proxemics. We discuss current limitations, design choices, and possible extensions of the metrics.
Our next target is to improve the grounding of $R$ and $E$ by evaluating trajectories of human interactions.
%
%
\bibliographystyle{IEEEtran}
\bibliography{root}

@article{gao_evaluation_2022,
	title={Evaluation of {S}ocially-{A}ware {R}obot {N}avigation},
	author={Gao, Yuxiang and Huang, Chien-Ming},
	journal={Frontiers in Robotics and AI},
	volume={8},
	pages={721317},
	year={2022},
	publisher={Frontiers Media SA}
}

@article{francis_principles_2025,
	title={Principles and {G}uidelines for {E}valuating {S}ocial {R}obot {N}avigation {A}lgorithms},
	author={Francis, Anthony and P{\'e}rez-d’Arpino, Claudia and Li, Chengshu and Xia, Fei and Alahi, Alexandre and Alami, Rachid and others},
	journal={ACM Transactions on Human-Robot Interaction},
	volume={14},
	number={2},
	pages={1--65},
	year={2025},
	publisher={ACM New York, NY}
}

@inproceedings{wang_metrics_2022,
	title={Metrics for {E}valuating {S}ocial {C}onformity of {C}rowd {N}avigation {A}lgorithms},
	author={Wang, Junxian and Chan, Wesley P and Carreno-Medrano, Pamela and Cosgun, Akansel and Croft, Elizabeth},
	booktitle={2022 IEEE International Conference on Advanced Robotics and Its Social Impacts (ARSO)},
	pages={1--6},
	year={2022},
	organization={IEEE}
}

@article{mirsky_conflict_2024,
	title={Conflict avoidance in social navigation—a survey},
	author={Mirsky, Reuth and Xiao, Xuesu and Hart, Justin and Stone, Peter},
	journal={ACM Transactions on Human-Robot Interaction},
	volume={13},
	number={1},
	pages={1--36},
	year={2024},
	publisher={ACM New York, NY}
}

@book{hall_hidden_1966,
	title={The hidden dimension},
	author={Hall, Edward T},
	volume={609},
	year={1966},
	publisher={Anchor}
}

@article{lee_visual_1977,
	title={Visual control of locomotion},
	author={Lee, David N and Lishman, Roly},
	journal={Scandinavian Journal of Psychology},
	volume={18},
	number={1},
	pages={224--230},
	year={1977},
	publisher={Wiley Online Library}
}

@inproceedings{eggert_predictive_2014,
	title={Predictive {R}isk {E}stimation for intelligent {ADAS} {F}unctions},
	author={Eggert, Julian},
	booktitle={17th International IEEE Conference on Intelligent Transportation Systems (ITSC)},
	pages={711--718},
	year={2014},
	organization={IEEE}
}

@inproceedings{wenzel_spotting_2025,
	location = {Atlanta, {GA}, {USA}},
	title = {Spotting the {U}nfriendly {R}obot - {T}owards better {M}etrics for {I}nteractions},
	booktitle = {{W}orkshop on {S}ocial {R}obot {N}avigation, {ICRA}},
	author = {{Wenzel, Raphael} and Probst, Malte},
	year= {2025}
}

@article{helbing_social_1995,
	title = {Social {F}orce {M}odel for {P}edestrian {D}ynamics},
  author={Helbing, Dirk and Molnar, Peter},
  journal={Physical review E},
  volume={51},
  number={5},
  pages={4282},
  year={1995},
  publisher={APS}
}

@article{fox_dynamic_1997,
  title={The {D}ynamic {W}indow {A}pproach to {C}ollision {A}voidance},
  author={Fox, Dieter and Burgard, Wolfram and Thrun, Sebastian},
  journal={IEEE Robotics \& Automation Magazine},
  volume={4},
  number={1},
  pages={23--33},
  year={1997},
  publisher={IEEE}
}

@inproceedings{george_feasible_2023,
	title={Feasible action-space reduction as a metric of causal responsibility in multi-agent spatial interactions},
	author={George, Ashwin and Siebert, Luciano Cavalcante and Abbink, David and Zgonnikov, Arkady},
	booktitle="26th European Conference on Artificial Intelligence, ECAI 2023",
	year={2023},
	pages = "819--826",
}

@article{babel_findings_2022,
	title={Findings from a qualitative field study with an autonomous robot in public: exploration of user reactions and conflicts},
	author={Babel, Franziska and Kraus, Johannes and Baumann, Martin},
	journal={International Journal of Social Robotics},
	volume={14},
	number={7},
	pages={1625--1655},
	year={2022},
	publisher={Springer}
}

@inproceedings{reeves_opening_2025,
	title={Opening Up Human-Robot Collaboration},
	author={Reeves, Stuart and Pelikan, Hannah and Cantarutti, Marina},
	booktitle={Proceedings of the ACM on Human-Computer Interaction},
	year={2025},
}

@incollection{vincent2011structured,
  title={A structured taxonomy of responsibility concepts},
  author={Vincent, Nicole A},
  booktitle={Moral responsibility: Beyond free will and determinism},
  pages={15--35},
  year={2011},
  publisher={Springer}
}

@inproceedings{remy2025learning,
  title={Learning responsibility allocations for multi-agent interactions: A differentiable optimization approach with control barrier functions},
  author={Remy, Isaac and Fridovich-Keil, David and Leung, Karen},
  booktitle={2025 American Control Conference (ACC)},
  pages={3213--3220},
  year={2025},
  organization={IEEE}
}

@article{shalev2017formal,
  title={On a formal model of safe and scalable self-driving cars},
  author={Shalev-Shwartz, Shai and Shammah, Shaked and Shashua, Amnon},
  journal={arXiv preprint arXiv:1708.06374},
  year={2017}
}

\end{document}